\documentclass[10pt,twocolumn,letterpaper]{article}

\usepackage[pagenumbers]{wacv} 

\definecolor{wacvblue}{rgb}{0.21,0.49,0.74}
\usepackage[pagebackref,breaklinks,colorlinks,allcolors=wacvblue]{hyperref}

\usepackage[accsupp]{axessibility}  
\usepackage{graphicx}
\usepackage{booktabs}
\usepackage{tabularx}
\usepackage{multirow}
\usepackage{relsize}

\newcommand{\rev}[1]{{#1}}

\def\wacvPaperID{510} 
\def\confName{WACV}
\def\confYear{2027}

\title{StepControl: Long-Horizon Human Motion Generation\\ from Foot-Contact Schedules}

\author{Hojun Jang$^1$ \and Jinseok Bae$^1$ \and Donggeun Lim$^1$ \and Taesun Kim$^1$ \and Seungbin Jeong$^2$ \and Younghwan Lee$^1$ \and Yong Seok Kim$^1$ \and Young Min Kim$^{1,2,3}$\thanks{Corresponding author} \\
\and
$^1$ Dept. of Electrical and Computer Engineering, Seoul National University \\
$^2$ Interdisciplinary Program in Artificial Intelligence, Seoul National University \\
$^3$ Institute of New Media and Communications, Seoul National University \\ \\
{\tt \relsize{-0.5} \href{https://hojunjang17.github.io/StepControl}{https://hojunjang17.github.io/StepControl}}
}

\begin{document}
\addtocounter{footnote}{1}
\maketitle
\begin{abstract}
Controllable human motion generation has largely focused on specifying what action should occur or where the body should move. 
We study a complementary question: can full-body motion be controlled by specifying only when each foot contacts the ground?
We introduce StepControl, an autoregressive diffusion model conditioned on a future binary left/right foot-contact schedule, without requiring target root and body poses, or semantic prompts.
StepControl represents contact as explicit temporal events and injects them through dedicated cross-attention with classifier-free guidance.
We further introduce a stochastic mixture noise schedule and warmup strategy for stable long-horizon autoregressive generation.
\rev{Experiments show that StepControl follows prescribed contact timing more accurately than adapting existing spatially guided motion generators to contact constraints.}
It maintains effective contact control over long continuous rollouts and responds to edited schedules such as cadence changes, stop-and-go pauses, limping gaits, and bilateral jumping.
Our results show that minimal foot-contact schedules provide an effective and intuitive temporal interface for controllable motion generation.
\end{abstract}
    
\section{Introduction}

\begin{figure}[t]
\begin{center}

\includegraphics[width=\linewidth]{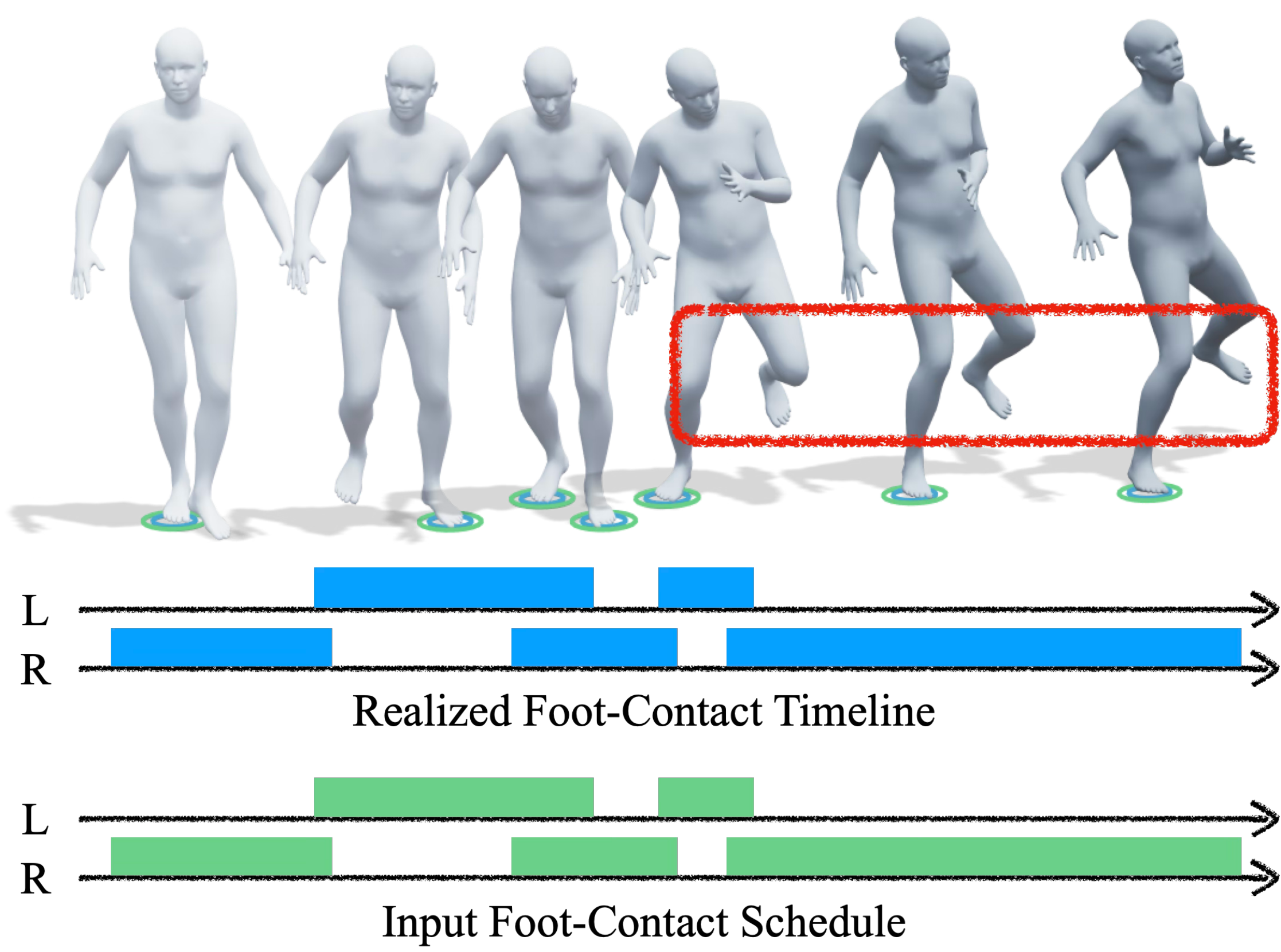}
\caption{
StepControl generates full-body motion from foot-contact schedules without future foot positions or root trajectories.
Beyond reproducing ordinary walking cadence, StepControl follows atypical asymmetric schedules.
}
\label{fig:teaser}
\end{center}
\vspace{-2em}
\end{figure}

Recent advances in motion generation demonstrate improved diversity and fidelity, expressing rich semantics and styles of motion~\cite{snapmogen,scalemogen,unsup_styles,kimodo,ardy}.
These methods largely learn from skeletal motion data and often focus on the human motion itself without explicitly modeling interactions with the environment.
However, many motions are strongly structured by physical contacts with the surrounding world.
Physics-based character animation explicitly models such interactions through dynamics and contact~\cite{amp,deepmimic,maskedmimic}, but typically requires a physics simulator with task-specific control policies and objectives.
We instead investigate whether contact timing itself can serve as a simple and intuitive interface for conditioning a generative motion model.

In this work, we propose to first define the core of the motion from the contact timing only, and use the power of a generative model to decide the global trajectory, speed, stride pattern, and expressive body dynamics.
The contact information is sparse and simple, but can allow editing cadence, stop-and-go behavior, bilateral jumps, and asymmetric stepping patterns, without specifying body poses.
In the real world, a possible source of contact information may be diverse multi-modal sensors, such as inertial, pressure, tactile sensors, or sound.
The contact schedule naturally complements motion estimation when the visual information is not available due to occluding objects or when a human moves out of the viewing frustum.

Compared to semantic context or kinematic priors, the minimal input of binary contact information is extremely low-dimensional.
Therefore, naively concatenating the input often results in ignoring the contact information~\cite{guo2022humanml3d,snapmogen}.
Incorporating such information as post-processing is also not desirable, creating unnatural artifacts, such as foot pinning and stiff steps, that severely degrade the quality of motion~\cite{footskate, reducing_footskate, LoBSTr}.
Sparse temporal events are also difficult to preserve under repeated autoregressive rollouts~\cite{dart,ardy}, as small prediction errors can accumulate across rollout windows~\cite{diffusionforcing,rpm} and eventually cause phase mismatches, popping artifacts, or gait collapse.

We introduce StepControl, an autoregressive motion diffusion model conditioned on a foot-contact schedule.
A key challenge is that binary contact schedules are sparse and contain sharp transition events that can be easily overlooked by a motion denoiser. 
We therefore explicitly represent contact states, strike onsets, lift releases, and signed transitions as event-aware features, which condition a Transformer-based diffusion denoiser through dedicated cross-attention.
For autoregressive synthesis, we adopt frame-wise diffusion levels~\cite{zhang2024tedi,diffusionforcing,egoforce}, maintaining clean history together with progressively refined future states in a shared temporal window.
Since the corresponding future contact schedule is available within this window, the model can anticipate upcoming contact transitions and adapt the preceding motion accordingly, rather than forcing a sudden kinematic response at the moment of contact.
Progressive refinement further maintains smooth continuity as generated frames are committed back into the motion history.

Our contributions are summarized as follows:
\begin{itemize}
\item \textbf{Contact-Timing Control:} We introduce a motion generation pipeline conditioned on a minimal foot-contact schedule, demonstrating that two binary channels can direct rich full-body locomotion without spatial trajectories or semantic prompts.
\item \rev{\textbf{Event-Aware Contact Conditioning:} We show that learning an event-aware conditional motion prior achieves superior contact adherence compared to separately enforcing contact as spatial guidance.}
\item \rev{\textbf{Long-Horizon Controllability \& Interactive Editing:} We provide empirical evidence that effective temporal control persists over continuous 20\,s autoregressive rollout and responds causally to user-guided schedule edits.}
\end{itemize}

\section{Related Works}

\subsection{Autoregressive Motion Diffusion}
Diffusion models demonstrate their capability to generate diverse and complex human motions.
Early methods generate full-body motion directly in the motion space~\cite{mdm,motiondiffuse}, while later methods improve fidelity and efficiency through retrieval augmentation~\cite{remodiffuse} or compact latent spaces~\cite{mld,motionlcm}.
Most of these methods generate fixed-length clips from text prompts or action labels.
In contrast, we use frame-level foot contacts as a temporal control signal.

Long-horizon motion generation extends beyond fixed-length synthesis by recurrently conditioning on previously generated motion.
Several diffusion-based approaches adopt causal or autoregressive generation to produce continuous motion streams~\cite{dart,motionstreamer,amdm,ardy}.
Related works such as TEDi~\cite{zhang2024tedi}, Diffusion Forcing~\cite{diffusionforcing}, and EgoForce~\cite{egoforce} further use temporally varying diffusion levels to progressively refine motion over time.
We adopt this progressive formulation for contact-conditioned generation, allowing upcoming foot-contact events to influence the preceding motion.

\subsection{Controllable Motion Generation}
Human motion can be controlled with semantic signals, such as text and action labels, or with spatial signals, such as root trajectories, keyframes, and joint paths.
GMD~\cite{gmd} guides motion with a global trajectory, while OmniControl~\cite{omnicontrol} and PriorMDM~\cite{shafir2023priormdm} support spatial constraints on selected joints.
Kimodo~\cite{kimodo}, DART~\cite{dart}, and ARDY~\cite{ardy} further support keyframe, trajectory, or goal-based control.
\rev{In contrast to these spatial controls, a foot-contact schedule provides direct temporal control over locomotion by specifying when each foot should be planted or lifted, without fixing its position or the future root trajectory.}

Temporal conditioning has also been explored in music-driven dance generation.
Existing methods use acoustic and rhythmic features to coordinate motion with music~\cite{fact,listen_denoise_action,edge}, and Beat-It~\cite{beatit} explicitly separates beat events from continuous music features.
However, a musical beat does not directly indicate a foot strike, identify the active foot, or specify the length of a stance phase.
We use explicit left- and right-foot contact labels, which provide direct control over foot-strike timing, stance duration, cadence, and pauses.

\begin{figure*}[t]
\begin{center}

\includegraphics[width=\linewidth]{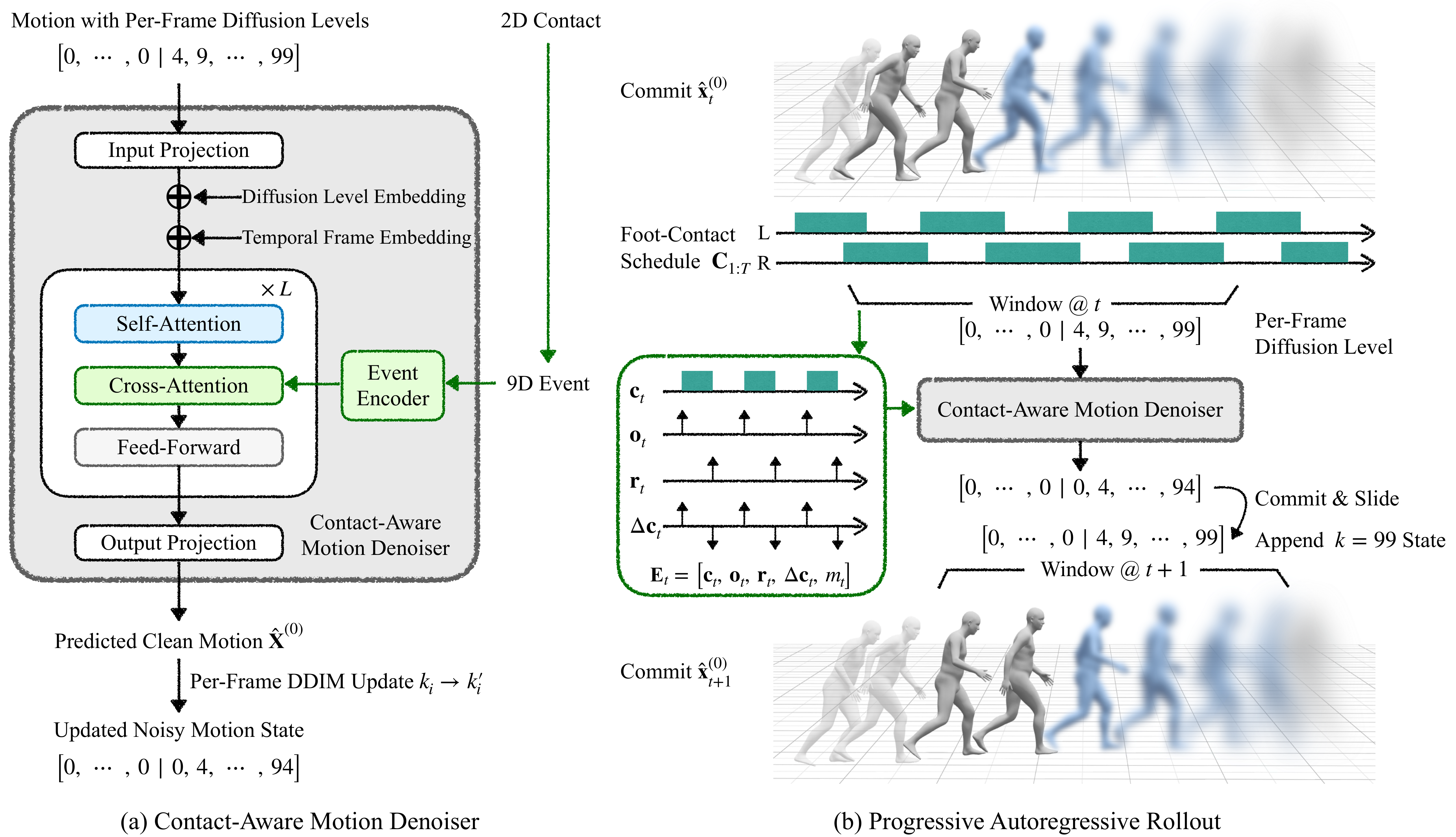}
\caption{
    \textbf{Overview of StepControl.} 
(a) Motion tokens with per-frame diffusion levels and frame embeddings attend to event condition tokens via dedicated per-layer cross-attention, predicting clean motion $\hat{\mathbf{X}}^{(0)}$ for continuous long-horizon generation.
(b) Within each sliding window, 2D foot-contact signals $\mathbf{C}$ are expanded into 9D event-aware representations $\mathbf{E}$. Conditioned on $\mathbf{E}$, the denoiser steps down noisy future latents via per-frame DDIM updates ($k_i \to k_i'$), commits the earliest finalized frame $\hat{\mathbf{x}}_t^{(0)}$ to clean history, and appends a new terminal noise state ($k=99$) as the window slides forward.
}
\label{fig:method}
\end{center}
\vspace{-2em}
\end{figure*}

\subsection{Foot Contact in Motion Generation}

Foot contact has commonly been incorporated into kinematic motion generation as an auxiliary cue for improving foot-ground consistency.
Binary contact labels are included in motion representations~\cite{guo2022humanml3d}, while training objectives penalize foot velocity during predicted contact to reduce artifacts such as foot skating and floating~\cite{mdm,edge,humor}.
Inverse kinematics and other kinematic corrections are also applied after motion generation to enforce plausible foot placement~\cite{footskate,reducing_footskate,underpressure,LoBSTr}.
In these approaches, contact primarily serves to regularize or correct the generated motion rather than to specify how the future motion should unfold.
In contrast, StepControl treats a prescribed future foot-contact schedule as the conditioning signal that directly drives motion generation.

Foot-ground signals have also been used as observations for full-body motion reconstruction.
Jang~\etal~\cite{audio_aided_tracking} combine footstep sounds with sparse inertial sensors; Step2Motion~\cite{step2motion} reconstructs locomotion from pressure and inertial measurements of sensing insoles; and GRIP~\cite{grip} combines plantar pressure with sparse body-mounted IMUs for physics-based motion reconstruction.
Our setting differs in that we use only a binary left/right contact schedule, without continuous sensor measurement, to condition the generation of full-body motion.

\section{StepControl}

Given an observed motion prefix $\mathbf{X}_{1:S}^{(0)}$ and a binary left/right foot-contact schedule $\mathbf{C}_{1:T}$, our goal is to model the conditional distribution of the future motion $\mathbf{X}_{S+1:T}^{(0)}$:
\begin{equation}
    p_\theta(\mathbf{X}_{S+1:T}^{(0)} \mid \mathbf{X}_{1:S}^{(0)}, \mathbf{C}_{1:T}),
\end{equation}
without requiring future 3D root trajectories, target keyframe poses, or high-level semantic prompts.
We adopt the 263-dimensional motion representation proposed in HumanML3D~\cite{guo2022humanml3d}, which jointly describes root movement, body pose, joint motion, and foot contact.
\rev{Throughout, the denoiser operates on a sliding window of $W=S+F$ frames, consisting of $S$ clean history frames and $F$ noisy future frames.}

We first describe how the binary foot-contact schedule is injected into the motion denoiser through dedicated cross-attention in~\cref{sec:contact_conditioning}.
Then, we describe the training objectives used to preserve motion quality and foot-ground consistency in~\cref{sec:loss_formulation}.
Finally, in~\cref{sec:rollout_and_recipe}, we present the progressive autoregressive rollout and its deployment-aware noise schedule for long-horizon generation.

\subsection{Event-Aware Contact Conditioning}
\label{sec:contact_conditioning}
StepControl represents the foot-contact schedule as a separate sequence of event-aware condition tokens and injects them into each Transformer layer through cross-attention, as illustrated in~\cref{fig:method}(a).
A binary contact signal $\mathbf{c}_t = [c_{t,\text{left}}, c_{t,\text{right}}] \in \{0, 1\}^2$ specifies whether each foot is in stance ($1$) or swing ($0$). 
While the complete contact sequence implicitly contains transition information, a raw contact state does not distinguish a newly occurring touchdown or lift-off from the interior of a stance or swing interval.
The denoiser must therefore infer such event boundaries from neighboring contact states.
To expose these temporally localized transitions directly, we expand $\mathbf{c}_t$ into a 9-dimensional event-aware representation $\mathbf{E}_t \in \mathbb{R}^9$:
\begin{equation}
    \mathbf{E}_t = \big[ \mathbf{c}_t, \; \mathbf{o}_t, \; \mathbf{r}_t, \; \Delta \mathbf{c}_t, \; m_t \big],
\end{equation}
where:
\begin{itemize}
    \item $\mathbf{o}_t = \max(\mathbf{c}_t - \mathbf{c}_{t-1}, \mathbf{0}) \in \{0, 1\}^2$ explicitly captures foot strike (\textit{onset}) events,
    \item $\mathbf{r}_t = \max(\mathbf{c}_{t-1} - \mathbf{c}_t, \mathbf{0}) \in \{0, 1\}^2$ captures foot lift (\textit{release}) events,
    \item $\Delta \mathbf{c}_t = \mathbf{c}_t - \mathbf{c}_{t-1} \in \{-1, 0, 1\}^2$ encodes signed temporal state transitions,
    \item $m_t \in \{0, 1\}$ is a binary observation validity mask.
\end{itemize}
Although the event channels are deterministically derived from the binary schedule, they make the timing and direction of contact transitions directly available to the motion denoiser, rather than requiring them to be inferred implicitly from neighboring contact states.

\paragraph{Per-Layer Contact Cross-Attention}
Rather than early concatenation, we inject the contact schedule into the diffusion Transformer via dedicated cross-attention layers operating on each local window of length $W$. 
Within a window, the local event-aware contact sequence $\mathbf{E}_{1:W} \in \mathbb{R}^{W \times 9}$ is linearly projected into contact condition tokens $\mathbf{Z} \in \mathbb{R}^{W \times d}$. 
In each Transformer layer $l$, local motion tokens $\mathbf{H}^{(l)} \in \mathbb{R}^{W \times d}$ perform multi-head cross-attention over $\mathbf{Z}$:
\begin{equation}
    \text{CrossAttn}(\mathbf{H}^{(l)}, \mathbf{Z}) = \text{softmax}\left(\frac{\mathbf{Q}_H \mathbf{K}_Z^\top}{\sqrt{d_k}}\right) \mathbf{V}_Z,
\end{equation}
where $\mathbf{Q}_H = \mathbf{H}^{(l)} \mathbf{W}_Q$, $\mathbf{K}_Z = \mathbf{Z} \mathbf{W}_K$, and $\mathbf{V}_Z = \mathbf{Z} \mathbf{W}_V$. 
This provides a direct, unhindered gradient pathway from contact timing to all levels of motion feature abstraction within the active generative window.

\paragraph{Paired Classifier-Free Guidance}
To provide flexible control over contact adherence at inference time, we train the denoiser using classifier-free guidance (CFG)~\cite{cfg}. 
Following paired-batch conditional training paradigms~\cite{infinity,mdm}, each batch sample is duplicated during training: one branch receives the local contact condition $\mathbf{E}_{1:W}$, while the paired branch receives a null condition $\mathbf{E}_{\emptyset} = \mathbf{0}$ sharing identical input motion noise. 
This coupled formulation eliminates gradient variance between conditional and unconditional estimates and ensures balanced score extrapolation. 
Let $\hat{\mathbf{X}}_{\text{cond}}^{(0)}$ and $\hat{\mathbf{X}}_{\text{uncond}}^{(0)}$ denote the clean motion predictions conditioned on $\mathbf{E}_{1:W}$ and $\mathbf{E}_{\emptyset}$, respectively. 
At inference time, the final denoised estimate is steered with guidance scale $s \ge 1.0$:
\begin{equation}
    \hat{\mathbf{X}}_{1:W}^{(0)} = \hat{\mathbf{X}}_{\text{uncond}}^{(0)} + s \cdot \big( \hat{\mathbf{X}}_{\text{cond}}^{(0)} - \hat{\mathbf{X}}_{\text{uncond}}^{(0)} \big).
\end{equation}

\subsection{Training Objectives}
\label{sec:loss_formulation}
The overall training objective combines future motion reconstruction alongside geometric and auxiliary foot-ground supervision. Every constituent loss term is evaluated symmetrically across both the conditional and null-conditioned CFG branches ($\mathcal{L}_* = \mathcal{L}_*^{\text{cond}} + \mathcal{L}_*^{\text{uncond}}$):
\begin{equation}
    \mathcal{L} = \mathcal{L}_{\text{mse}}
    + \lambda_{\text{vel}}\mathcal{L}_{\text{stance}}
    + \lambda_{\text{h}}\mathcal{L}_{\text{height}}
    + \lambda_{\text{c}}\mathcal{L}_{\text{contact4}}.
\end{equation}

\paragraph{Motion Reconstruction Loss}
The observed clean motion prefix $\mathbf{X}_{1:S}^{(0)}$ is provided directly as clean historical context at noise level $k=0$. 
The motion reconstruction loss $\mathcal{L}_{\text{mot}}$ serves as the primary diffusion training objective, supervising the model's clean motion prediction $\hat{\mathbf{X}}_{S+1:W}^{(0)}$ over the full standardized 263-dimensional feature space across all future frames ($t \in [S+1, W]$):
\begin{equation}
    \mathcal{L}_{\text{mse}} =
    \operatorname{MSE}(\hat{\mathbf{X}}_{S+1:W}^{(0)},\mathbf{X}_{S+1:W}^{(0)}).
\end{equation}
By supervising the complete feature vector directly, $\mathcal{L}_{\text{mse}}$ trains the denoiser to jointly model coupled full-body kinematics, root velocities, and joint rotations without arbitrary dimensional weighting.

\paragraph{Stance Foot Velocity Loss}
We penalize recovered 3D foot displacement when the target four-channel contact label $c^{\text{hml}}_{t,j}$ indicates stance:
\begin{equation}
    \mathcal{L}_{\text{stance}} =
    \frac{\sum_{t,j} c^{\text{hml}}_{t,j}
    \|\hat{\mathbf{J}}_{t+1,j}-\hat{\mathbf{J}}_{t,j}\|_2^2}
    {\sum_{t,j}c^{\text{hml}}_{t,j}},
\end{equation}
where $j \in \mathcal{F}$ indexes the left/right ankle and foot joints used by HumanML3D contact extraction.

\paragraph{Foot Height Loss}
To preserve the vertical geometry of both stance and swing phases, we match recovered foot heights to the ground-truth motion over every future frame:
\begin{equation}
    \mathcal{L}_{\text{height}} =
    \frac{1}{|\mathcal{T}_{f}||\mathcal{F}|}
    \sum_{t\in\mathcal{T}_{f}}\sum_{j\in\mathcal{F}}
    \left(\hat{y}_{t,j}-y_{t,j}\right)^2.
\end{equation}

\paragraph{Auxiliary Contact Bit Loss}
Within the full 263-dimensional motion vector $\mathbf{x}_t^{(0)}$, the 4 binary contact channels account for only $\approx 1.5\%$ of the total MSE reconstruction loss $\mathcal{L}_{\text{mse}}$, causing their gradient updates to be heavily diluted by continuous joint positions and 6D rotations. 
To prevent this dilution and enforce sharp discrete ground transitions, we apply an explicit auxiliary reconstruction loss directly on the 4 contact channels:
\begin{equation}
    \mathcal{L}_{\text{contact4}} =
    \operatorname{MSE}(\hat{\mathbf{c}}^{\text{hml}},\mathbf{c}^{\text{hml}}).
\end{equation}

\subsection{Progressive Autoregressive Rollout}
\label{sec:rollout_and_recipe}

StepControl generates long-horizon motion using a sliding window of $W=30$ frames, consisting of $S=10$ clean history frames and $F=20$ progressively refined future frames as shown in~\cref{fig:method}(b).
Each frame is associated with its own diffusion level $k \in \{0,\ldots,99\}$, where $k=0$ denotes a clean state and larger values indicate increasing uncertainty.

\paragraph{Progressive Noise Schedule}
During autoregressive rollout, future frames are assigned increasing diffusion levels according to their temporal distance from the committed history.
Specifically, we use the fixed rollout schedule
\begin{equation}
    \mathbf{k}_{\mathrm{roll}}
    =
    [4, 9, 14, \ldots, 99].
    \label{eq:rollout_schedule}
\end{equation}
Thus, the nearest future frame is close to being finalized, whereas frames farther into the future remain increasingly noisy.
Rather than independently regenerating an entire future segment at each autoregressive step, each future frame is progressively refined as it approaches the current time.
\rev{Since every frame is generated while observing the contact schedule over a constant future horizon, the model can prepare for upcoming touchdowns and lift-offs at each step of the rollout.
We compare this against chunk-wise rollout in the Supplementary Material.}

\paragraph{Mixture Noise Training}
The rollout schedule in Eq.~\eqref{eq:rollout_schedule} covers the noise configurations repeatedly encountered during steady generation.
However, it does not cover all intermediate configurations traversed when the first future window is initialized from Gaussian noise.
To support both regimes, we train with a mixture of rollout-aligned and independently sampled per-frame noise levels.
For each training window, the future noise-index vector is sampled as
\begin{equation}
\mathbf{k}_{\mathrm{future}}
=
\begin{cases}
    \mathbf{k}_{\mathrm{roll}},
        & \text{with probability } 0.70, \\[2mm]
    [k_{S+1},\ldots,k_W],
        & \text{with probability } 0.30,
\end{cases}
\label{eq:mixture_noise}
\end{equation}
where, in the second case,
\begin{equation}
    k_t \overset{\mathrm{i.i.d.}}{\sim}
    \mathrm{Unif}\{0,\ldots,99\}.
\end{equation}
The rollout-aligned samples specialize the denoiser for the structured noise configurations used repeatedly during steady autoregressive generation, while the IID samples broaden its support to the intermediate configurations encountered during initialization.

\paragraph{Initialization and Steady Rollout}
At the beginning of generation, only the clean motion prefix is available, and the future states have not yet been placed on $\mathbf{k}_{\mathrm{roll}}$.
We therefore initialize all $F$ future frames at the maximum noise level, $k=99$, and progressively denoise them to their assigned rollout levels while keeping the observed history fixed.
This multi-step warmup avoids committing the first generated frame directly from a fully noisy future window.
The IID component of Eq.~\eqref{eq:mixture_noise} exposes the denoiser during training to the intermediate noise configurations traversed by this initialization procedure.

Once the progressive noise schedule has been initialized, StepControl enters steady autoregressive rollout.
Given the current per-frame noisy window, the denoiser predicts a clean motion window $\hat{\mathbf{X}}_0$.
A per-frame DDIM update then refines the future states according to
\begin{equation}
    [4,9,14,\ldots,99]
    \;\longrightarrow\;
    [0,4,9,\ldots,94].
    \label{eq:rollout_ddim_update}
\end{equation}
The first future frame has now reached $k=0$ and is committed to the generated sequence.
The motion and contact windows are then advanced by one frame, the committed prediction becomes part of the clean history, and a new Gaussian state
\begin{equation}
    \mathbf{x}^{(99)}_{\mathrm{new}}
    \sim \mathcal{N}(\mathbf{0},\mathbf{I})
\end{equation}
is appended at the end of the future window.
After this shift, the future frames again follow $\mathbf{k}_{\mathrm{roll}}$, and the same procedure is repeated.

Consequently, each newly appended frame is refined over successive autoregressive steps as
\begin{equation}
    \mathbf{x}^{(99)}
    \rightarrow
    \mathbf{x}^{(94)}
    \rightarrow
    \cdots
    \rightarrow
    \mathbf{x}^{(4)}
    \rightarrow
    \hat{\mathbf{x}}^{(0)}.
    \label{eq:progressive_refinement}
\end{equation}
A future frame therefore receives multiple refinement steps, each time conditioned on the latest committed motion history and the corresponding foot-contact schedule.
This repeated refinement provides temporal look-ahead into upcoming contact events, allowing the model to adapt the preceding motion before the corresponding contact transition occurs.
Repeating this commit--slide--append procedure enables arbitrary-length, contact-conditioned autoregressive motion generation.

\section{Experiments}
\begin{table*}[t]\caption{\textbf{Quantitative comparison on autoregressive motion generation across extended horizons.} We evaluate contact synchronization and progressive rollout stability against autoregressive baselines (DART~\cite{dart}, ARDY~\cite{ardy}) across LAFAN1~\cite{lafan1} test sequences.}
\centering
\small
\resizebox{0.95\linewidth}{!}{
\begin{tabular}{llccccc}
\toprule
\textbf{Rollout Horizon} & \textbf{Method} & \textbf{Bal. Acc. (\%)} $\uparrow$ & \textbf{Tr. F1 (\%)} $\uparrow$ & \textbf{Slip (mm/f)} $\downarrow$ & \textbf{RA-LB (mm)} $\downarrow$ & \textbf{Mean Pen. (mm)} $\downarrow$ \\
\midrule
\multirow{4}{*}{\textbf{1-Second}}
 & DART & 74.34 & 15.26 & \textbf{7.84} & \textbf{109.97} & 0.30 \\
 & ARDY & 69.37 & 16.92 & 20.69 & 158.62 & 0.47 \\
 & \textbf{StepControl} & 79.90 & \textbf{36.15} & 17.63 & 144.25 & \textbf{0.00} \\
 & StepControl (+Opt.) & \textbf{80.60} & 33.59 & 8.64 & 144.25 & \textbf{0.00} \\

\midrule
\multirow{4}{*}{\textbf{5-Second}}
 & DART & 69.25 & 16.52 & \textbf{3.28} & \textbf{180.68} & 0.31 \\
 & ARDY & 64.28 & 10.29 & 13.52 & 185.86 & 0.34 \\
 & \textbf{StepControl} & 79.38 & 39.94 & 15.54 & 181.64 & \textbf{0.00} \\
 & StepControl (+Opt.) & \textbf{80.17} & \textbf{39.96} & 8.23 & 181.64 & \textbf{0.00} \\

\midrule
\multirow{4}{*}{\textbf{10-Second}}
 & DART & 68.30 & 15.53 & \textbf{2.72} & 208.15 & 0.30 \\
 & ARDY & 67.07 & 10.91 & 12.34 & \textbf{186.80} & 0.28 \\
 & \textbf{StepControl} & 80.28 & \textbf{40.07} & 16.83 & 194.70 & \textbf{0.00} \\
 & StepControl (+Opt.) & \textbf{81.10} & 39.84 & 8.97 & 194.70 & \textbf{0.00} \\

\midrule
\multirow{4}{*}{\textbf{20-Second}}
 & DART & 69.15 & 14.60 & \textbf{2.41} & 222.43 & 0.16 \\
 & ARDY & 76.02 & 15.98 & 7.83 & \textbf{187.54} & 0.16 \\
 & \textbf{StepControl} & 82.39 & 42.27 & 14.62 & 193.35 & \textbf{0.00} \\
 & StepControl (+Opt.) & \textbf{83.04} & \textbf{43.79} & 7.75 & 193.35 & \textbf{0.00} \\
\bottomrule
\end{tabular}
}
\label{tab:long_horizon}
\end{table*}

\subsection{Experimental Setup}
\label{sec:exp_setup}

\paragraph{Datasets and Training Recipe}
\rev{We train in two stages. First, the denoiser is pretrained on the large-scale HumanML3D dataset~\cite{guo2022humanml3d} (263-D representation, 22 joints at 20\,fps) to learn broad kinematic dynamics. Then, it is fine-tuned on LAFAN1~\cite{lafan1} motion capture sequences (Subjects 1--3 for training, Subject 4 for validation), retargeted to the HumanML3D representation at 20\,fps, to sharpen footstep timing and cadence control.}
All evaluations are conducted on the held-out LAFAN1 Subject 5 test sequences.

\begin{figure}[t]
\begin{center}

\includegraphics[width=\linewidth]{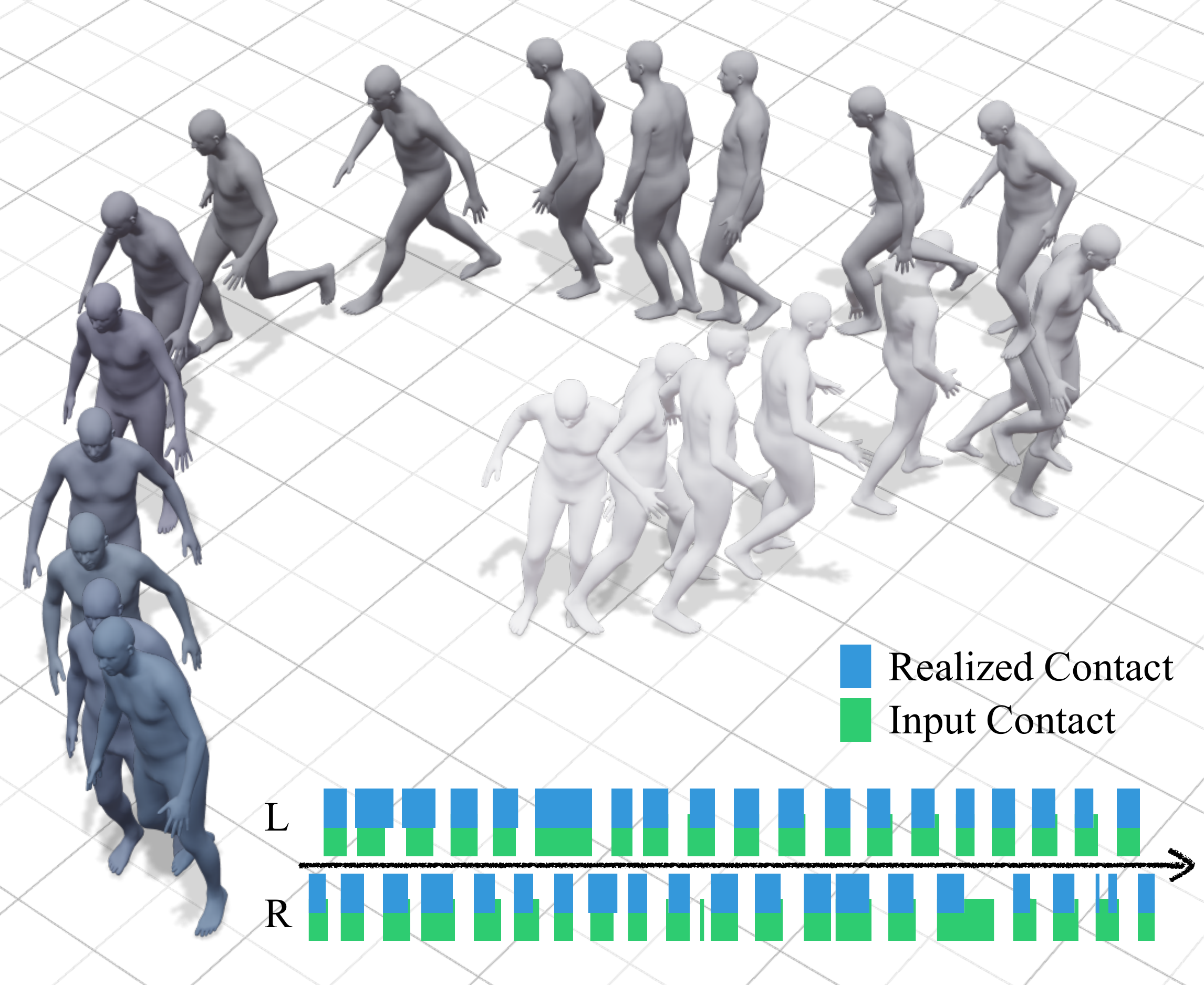}
\vspace{-2em}
\caption{
\textbf{20\,s Motion Generation Result of StepControl.} 
StepControl generates natural and consistent long-horizon motion from only foot-contact schedule.
}
\label{fig:long_rollout}
\end{center}
\vspace{-2em}
\end{figure}

\begin{figure*}[t]
\begin{center}

\includegraphics[width=0.95\linewidth]{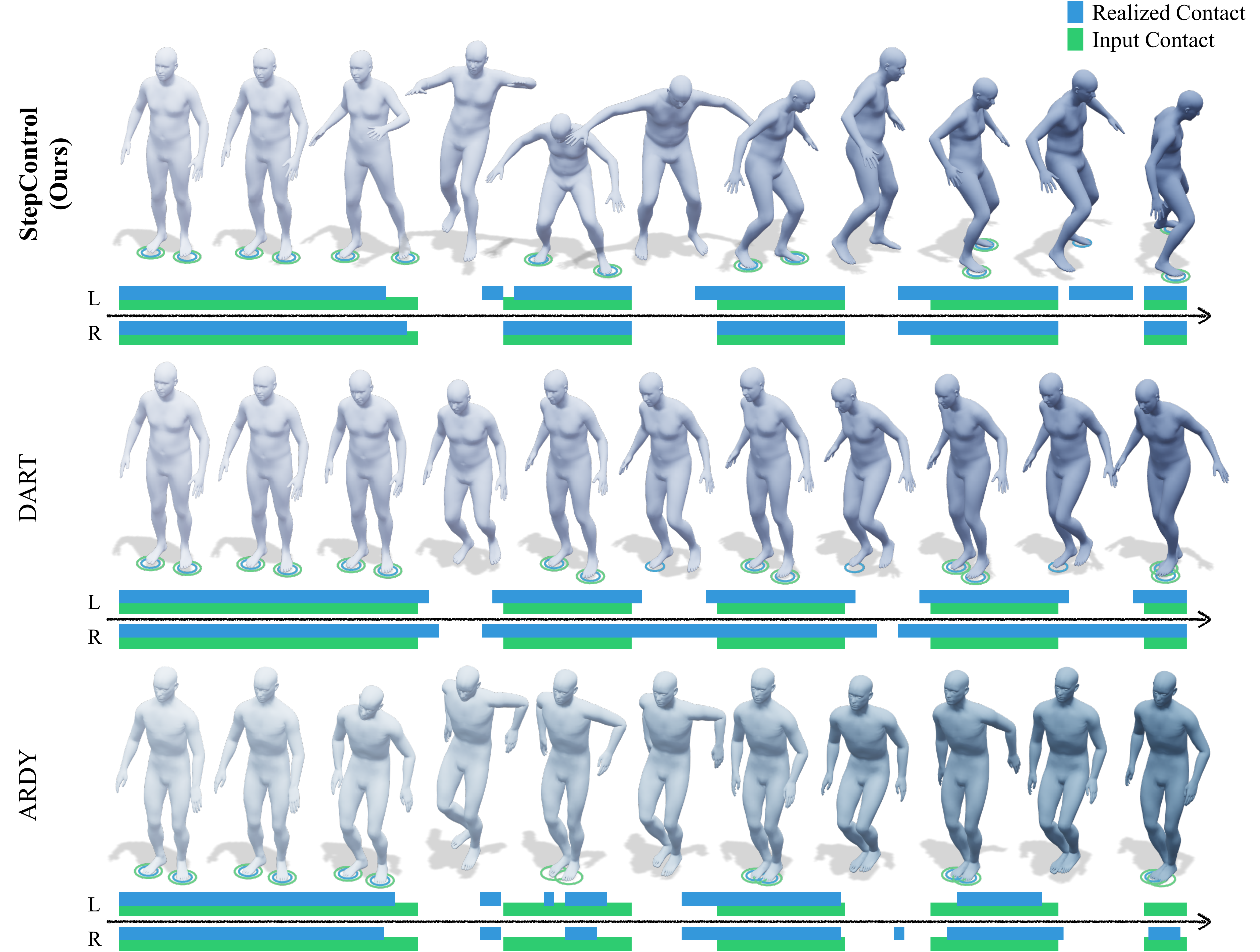}
\vspace{-0.0em}
\caption{
    \textbf{Qualitative comparison on contact schedule editing (bilateral jump).}
    StepControl better synchronizes simultaneous flight and landing phases, whereas DART~\cite{dart} fails to coordinate flight and ARDY~\cite{ardy} exhibits noisy contact timing and unstable poses.
    ARDY is rendered with its native body model.
}
\label{fig:edit}
\end{center}
\vspace{-2em}
\end{figure*}
\paragraph{Evaluation Metrics}

To evaluate motion quality and temporal synchronization, we measure:
\begin{itemize}
    \item \textbf{Balanced Contact Accuracy (\%):} The unweighted average of stance recall and swing recall, computed from 3D foot joint velocities and heights of decoded motions.
    \item \textbf{Contact Transition F1 (Tr. F1, \%):} While Balanced Accuracy is insensitive to temporal phase shifts, Tr. F1 quantifies frame-exact synchronization. We identify \textbf{Onset} ($0 \to 1$, foot touchdown) and \textbf{Release} ($1 \to 0$, foot lift-off) events, compute the F1-score of each against the target schedule, and report $\text{Tr. F1} = \frac{1}{2}(\text{F1}_{\text{onset}} + \text{F1}_{\text{release}})$.
    \item \textbf{Stance Foot Slip (mm/frame):} Mean horizontal displacement of foot joints during frames where the target schedule specifies stance.
    \item \textbf{Floor Penetration (mm):} Mean vertical distance of generated foot joints below the physical reference floor level.
    \item \textbf{Root-Aligned Lower-Body MPJPE (mm):} Mean per-joint position error over lower-body joints (hips, knees, ankles, and toes) after per-frame root translation and heading alignment, following~\cite{audio_aided_tracking}.
    \item \textbf{Boundary Jerk (mm/frame$^3$):} Mean root-relative joint jerk at the observed-to-generated transition averaged over non-root joints.
\end{itemize}

\paragraph{Baselines}

We compare StepControl against two autoregressive motion generation baselines, DART~\cite{dart} and ARDY~\cite{ardy}.
We adapt their spatial conditioning to contact schedules using foot-height constraints, while providing text prompts consistent with their training distributions.
Implementation and conditioning details are provided in the Supplementary Material.

\subsection{Long-Horizon Motion Generation}
\label{sec:long_horizon_generation}
To evaluate whether temporal contact control survives repeated autoregressive rollout, we generate continuous sequences of increasing horizons (1\,s, 5\,s, 10\,s, and 20\,s) on the LAFAN1~\cite{lafan1} test sequences using the progressive rollout of~\cref{sec:rollout_and_recipe}, and compare with the baselines under an identical evaluation protocol (\cref{tab:long_horizon}).
\rev{StepControl maintains stable contact timing across 20-second continuous rollouts with zero floor penetration.}
At 20\,s, raw StepControl reaches $42.27\%$ transition F1, compared with $14.60\%$ for DART and $15.98\%$ for ARDY, while retaining comparable lower-body error.

\begin{table}[t]
\caption{\textbf{Quantitative evaluation of interactive contact scheduling (5\,s horizon).} We evaluate responsiveness to temporal cadence scaling, gait transitions, and atypical gait synthesis against DART~\cite{dart} and ARDY~\cite{ardy} across LAFAN1~\cite{lafan1} test sequences.}
\vspace{-0.0em}
\centering
\small
\resizebox{\linewidth}{!}{
\begin{tabular}{llccc}
\toprule
\multirow{2}{*}{\textbf{Modification}} & \multirow{2}{*}{\textbf{Method}} & \multirow{2}{*}{\shortstack[c]{\textbf{Bal. Acc. $\uparrow$} \\ \textbf{(\%)}}} & \multirow{2}{*}{\shortstack[c]{\textbf{Tr. F1} $\uparrow$ \\ \textbf{(\%)}}} & \multirow{2}{*}{\shortstack[c]{\textbf{Slip} $\downarrow$ \\ \textbf{(mm/f)}}} \\
\\

\midrule
\multirow{3}{*}{{\textbf{Cadence Scaling}}}
 & DART & 67.91 & 9.13 & \textbf{3.44} \\
 & ARDY & 77.41 & 16.44 & 6.19 \\
 & \textbf{StepControl} & \textbf{76.11} & \textbf{33.57} & 16.49 \\
\midrule

\multirow{3}{*}{\textbf{Stop-and-Go}}
 & DART & 69.49 & 21.13 & \textbf{2.80} \\
 & ARDY & 78.88 & 16.81 & 4.05 \\
 & \textbf{StepControl} & \textbf{82.18} & \textbf{37.31} & 11.77 \\
\midrule

\multirow{3}{*}{\textbf{Limping Gait}}
 & DART & 66.22 & 6.56 & \textbf{6.15} \\
 & ARDY & 71.96 & 9.27 & 7.47 \\
 & \textbf{StepControl} & \textbf{75.55} & \textbf{31.25} & 27.46 \\
\midrule

\multirow{3}{*}{\textbf{Bilateral Jump}}
 & DART & 66.12 & 16.42 & \textbf{6.89} \\
 & ARDY & 76.69 & 12.80 & 8.48 \\
 & \textbf{StepControl} & \textbf{78.06} & \textbf{45.13} & 15.72 \\

\bottomrule
\end{tabular}
}
\vspace{-1em}
\label{tab:schedule_editing}
\end{table}

\begin{table*}[t]
\caption{\textbf{Ablation studies on the LAFAN1~\cite{lafan1} test benchmark.} We evaluate 5-second prefixes of the same 20-second evaluation segments used in the main benchmark. }
\vspace{-0.8em}
\centering
\small
\resizebox{0.95\linewidth}{!}{
\begin{tabular}{lccccc}
\toprule
\textbf{Ablation Configuration} & \textbf{Bal. Acc. (\%)} $\uparrow$ & \textbf{Tr. F1 (\%)} $\uparrow$ & \textbf{Slip (mm/f)} $\downarrow$ & \textbf{RA-LB (mm)} $\downarrow$ & \textbf{Bo. Jerk (mm/f$^3$)} $\downarrow$ \\
\midrule
\multicolumn{6}{c}{\textit{(a) Contact Feature Representation}} \\
Raw 2D Contact & 76.23 & 21.79 & 16.06 & 182.54 & 77.92 \\
Delta 5D Contact & \textbf{79.30} & 23.79 & 17.45 & \textbf{175.75} & 75.48 \\
\textbf{Event 9D Contact (Ours)} & \textbf{79.30} & \textbf{39.58} & \textbf{15.54} & 181.64 & \textbf{70.54} \\

\midrule
\multicolumn{6}{c}{\textit{(b) Training Dataset}} \\
LAFAN1 Only & \textbf{80.99} & 26.00 & 21.02 & \textbf{180.43} & 83.34 \\
HumanML3D Only & 77.19 & 38.08 & \textbf{10.79} & 187.62 & 82.54 \\
\textbf{HumanML3D+LAFAN1 (Ours)} & 79.30 & \textbf{39.58} & 15.54 & 181.64 & \textbf{70.54} \\
\midrule
\multicolumn{6}{c}{\textit{(c) Noise Scheduling Strategy}} \\
100\% Progressive Schedule & 75.94 & 37.30 & \textbf{15.22} & 210.89 & 161.77 \\
100\% IID Schedule & 73.91 & 31.52 & 21.68 & \textbf{174.33} & \textbf{69.80} \\
\textbf{Mixture Schedule (7:3) (Ours)} & \textbf{79.30} & \textbf{39.58} & 15.54 & 181.64 & 70.54 \\
\bottomrule
\end{tabular}
}
\vspace{-0.0em}
\label{tab:ablations}
\end{table*}

DART~\cite{dart} and ARDY~\cite{ardy} obtain lower stance slip because their test-time spatial guidance or post-processing explicitly anchors stance feet.
\rev{In contrast, StepControl generates motion directly from the diffusion prior without post-hoc spatial correction, as shown in~\cref{fig:long_rollout}.}
We additionally report \textbf{StepControl (+Opt.)}, which applies a simple planar stance-foot correction and improves 20-second slip from $14.62$ to $7.75\,\text{mm/frame}$ without changing body articulation or root-aligned pose error.

\subsection{Motion Control by Contact Scheduling}
\label{sec:motion_control_editing}
We evaluate whether StepControl follows edited schedules beyond the contact patterns commonly seen in training, using four schedule modifications:
\begin{itemize}
    \item \textbf{Cadence Scaling:} Compressing ($0.67\times$) or dilating ($1.5\times$) future stride duration.
    \item \textbf{Stop-and-Go:} Inserting a 2-second stationary pause amidst continuous walking.
    \item \textbf{Asymmetric Limping:} Imposing unbalanced left/right stance durations (16 vs. 6 frames).
    \item \textbf{Bilateral Jumping:} Alternating synchronized two-foot stance and flight phases.
\end{itemize}

StepControl consistently responds to the edited schedules without retraining and outperforms both baselines in contact timing as shown in~\cref{tab:schedule_editing}.
For bilateral jumping, StepControl synchronizes flight and landing phases more faithfully than the baselines, as shown in~\cref{fig:edit}.

\subsection{Ablation Studies}
\label{sec:ablations}
In ~\cref{tab:ablations}, we isolate each key design choice in StepControl.
\rev{Additional ablations and analyses, including a rollout strategy comparison and a breakdown by contact-transition rate, are in the Supplementary Material.}

\paragraph{Event-Aware Features}
\rev{In~\cref{tab:ablations}(a), we compare contact representations under identical training protocols.}
While raw 2D contact and signed transitions (Delta 5D) yield $21.79\%$ and $23.79\%$ transition F1, explicitly separating onset and release channels (Event 9D) raises transition F1 to $39.58\%$.
Event 9D also reduces stance slip and gives the lowest boundary jerk among the contact representations.

\paragraph{Training Dataset}
In~\cref{tab:ablations}(b), we compare LAFAN1-only scratch training, zero-shot transfer from HumanML3D~\cite{guo2022humanml3d}, and HumanML3D pretraining followed by LAFAN1 fine-tuning. 
\rev{Pretraining on HumanML3D improves contact timing over LAFAN1-only training, and fine-tuning on LAFAN1 further improves lower-body accuracy and boundary continuity over HumanML3D-only training.
The two-stage recipe thus achieves the best transition F1 and boundary jerk, and falls between the two single-dataset models on the remaining metrics.}

\paragraph{Noise Scheduling Strategy and Rollout Continuity}
A rigid deterministic progressive slope schedule achieves low stance slip but produces a pronounced discontinuity at the observed-to-generated boundary. 
IID schedule training reduces this discontinuity but weakens contact transitions. 
Our mixture schedule training obtains the highest transition F1 while keeping the boundary jerk close to IID training.
\rev{Rolling out the same model chunk by chunk instead of progressively lowers transition F1 and increases boundary jerk (Supplementary Material).}

\section{Conclusion}

We presented StepControl, an autoregressive diffusion framework that generates full-body motion from foot-contact schedules without predefined root trajectories or target poses.
StepControl expands sparse binary contacts into event-aware features and injects them through dedicated cross-attention.
Its mixture-noise training and progressive rollout support continuous long-horizon generation from a short observed prefix.
\rev{Our experiments show that the model follows prescribed and edited contact schedules over extended rollouts.}

StepControl jointly predicts root trajectories and foot joints within a single kinematic denoising process, without explicitly enforcing that stance feet remain stationary.
As a result, its raw output exhibits more foot slip than methods using spatial optimization or foot-locking post-processing.
A lightweight planar correction largely reduces this slip without changing body articulation, but remains a post-hoc rather than structural solution.
\rev{Some residual foot slip and occasional foot floating are inherent limits of the kinematic formulation, and unseen contact patterns such as single-foot hopping remain challenging.}
\rev{Future work could integrate physical contact constraints or differentiable inverse kinematics into diffusion sampling to reduce foot sliding without post-processing.}
Sparse temporal conditioning could extend beyond locomotion to broader interaction scenarios where visual observations are incomplete or unavailable.

\section*{Acknowledgement}
This work was partly supported by Institute of Information \& Communications Technology Planning \& Evaluation (IITP) grant funded by the Korea government(MSIT) [NO.RS-2021-II211343, Artificial Intelligence Graduate School Program (Seoul National Univ.)] and [RS-2025-25442338, AI star Fellowship Support Program(Seoul National Univ.)], and BK21 FOUR program of the Education and Research Program for Future ICT Pioneers, Seoul National University in 2026.

{
    \small
    \bibliographystyle{ieeenat_fullname}
    \bibliography{main}
}
\newpage
\clearpage
\appendix

\section*{Appendix}
\renewcommand{\thesection}{\Alph{section}}
\renewcommand{\thetable}{A\arabic{table}}
\renewcommand{\thefigure}{A\arabic{figure}}
\setcounter{section}{0}
\setcounter{table}{0}
\setcounter{figure}{0}

In this supplementary document, we provide comprehensive technical details and additional empirical analyses to support the main paper:
\begin{itemize}
    \item \textbf{\Cref{sec:supp_baselines}:} Detailed implementation, contact-adaptation protocols, and contact determination criteria for autoregressive baselines (DART~\cite{dart} and ARDY~\cite{ardy}).
    \item \textbf{\Cref{sec:supp_runtime}:} Computational runtime and throughput comparison across methods for 20-second continuous rollouts.
    \item \textbf{\Cref{sec:supp_ablations}:} Extended ablation studies on loss formulations and classifier-free guidance scales.
    \item \textbf{\Cref{sec:supp_foot_opt}:} Formulation and algorithm for the lightweight unconstrained planar stance-foot correction (\textbf{StepControl +Opt.}).
    \item \textbf{\Cref{sec:supp_arch_training}:} Detailed network architecture, training hyperparameters, and computational resources.
    \item \textbf{\Cref{sec:supp_metrics}:} Mathematical definitions and evaluation protocols for all quantitative metrics.
\end{itemize}

\section{Baseline Implementation Details}
\label{sec:supp_baselines}

Because existing autoregressive motion models are fundamentally designed for text or 3D spatial keyframe conditioning rather than sparse temporal contact schedules, we establish standardized adaptation protocols that allow both baselines to utilize their native architectures while conditioning on target contact schedules.

\subsection{DART Baseline Adaptation}
\label{sec:supp_dart}

\paragraph{Architecture and Pipeline}
DART~\cite{dart} generates continuous human motion using autoregressive latent motion primitives modeled by a Motion Variational Autoencoder (MVAE) operating at 20\,fps. Each primitive encompasses $H=2$ history frames and $F=8$ future frames (a 10-frame local temporal horizon). Generation proceeds by sequentially predicting latent primitive distributions conditioned on clean past history and natural language prompts.

\paragraph{Contact Conditioning via Test-Time Spatial Guidance}
Since DART lacks a dedicated temporal contact conditioning interface, we adapt it to follow target contact schedules through test-time gradient-based spatial guidance:
\begin{enumerate}
    \item \textbf{Matched Text Conditioning:} DART requires semantic text guidance during sampling. For each test sequence and editing mode, we supply matched text prompts consistent with its training distribution (e.g., \textit{``a person is walking forward''} for natural rollouts, \textit{``a person is moving with a fast cadence''} for fast editing, and \textit{``a person is repeatedly jumping with both feet''} for bilateral jumping).
    \item \textbf{Initial-Noise Contact Optimization:} At each autoregressive step, we optimize the diffusion initial noise for 50 iterations using the Adam optimizer ($\text{lr} = 0.05$), jointly over up to four primitives, and commit only the first primitive. The optimization objective enforces foot-ground contact constraints on decoded joints:
    \begin{equation}
    \begin{aligned}
        \mathcal{L}_{\text{opt}} = & \; w_{\text{st}}\mathcal{L}_{\text{stance\_height}} + w_{\text{sw}}\mathcal{L}_{\text{swing\_clearance}} + w_{\text{vel}}\mathcal{L}_{\text{stance\_vel}} \\
        & + w_{\text{fl}}\mathcal{L}_{\text{floor}} + w_{\text{jk}}\mathcal{L}_{\text{jerk}} + w_{\text{pr}}\mathcal{L}_{\text{prior}},
    \end{aligned}
    \end{equation}
    where:
    \begin{itemize}
        \item $\mathcal{L}_{\text{stance\_height}}$ penalizes deviations of the ankle joints from the stance height target during scheduled stance intervals ($w_{\text{st}} = 15.0$). The target is the calibrated stance ankle height, i.e., the median ankle height of each foot over the contact frames of the seed history, with $0.06\,\text{m}$ as fallback.
        \item $\mathcal{L}_{\text{swing\_clearance}}$ penalizes ankle heights below the calibrated stance ankle height plus $0.12\,\text{m}$ during scheduled flight intervals ($w_{\text{sw}} = 7.5$).
        \item $\mathcal{L}_{\text{stance\_vel}}$ penalizes horizontal displacement of stance feet ($w_{\text{vel}} = 10.0$).
        \item $\mathcal{L}_{\text{floor}}$ penalizes ankle and toe joint heights below the calibrated floor target minus $0.02\,\text{m}$ ($w_{\text{fl}} = 5.0$).
        \item $\mathcal{L}_{\text{jerk}}$ is a jerk smoothness term ($w_{\text{jk}} = 1.0$) and $\mathcal{L}_{\text{prior}}$ is a noise prior term ($w_{\text{pr}} = 0.1$).
    \end{itemize}
    \item \textbf{Prefix Alignment:} Global trajectory positions and root yaw headings are aligned to the shared initial 10-frame seed history ($\text{Error} = 0.00\,\text{mm}$), ensuring an identical evaluation starting frame.
\end{enumerate}

\subsection{ARDY Baseline Adaptation}
\label{sec:supp_ardy}

\paragraph{Architecture and Topology}
ARDY~\cite{ardy} is an autoregressive motion diffusion model with spatial inpainting capabilities. It operates natively on the 27-joint \textsc{CoreSkeleton27} topology~\cite{core}, which expands the 22-joint HumanML3D skeleton~\cite{guo2022humanml3d} by incorporating four discrete spinal segments and dedicated hand/wrist end-effectors.

\paragraph{Conditioning and Inpainting Recipe}
We deploy the official pretrained \textsc{ARDY-Core-RP-20FPS-Horizon40} checkpoint without retraining. It generates 40-frame sliding windows with a crop history length of 160 frames:
\begin{enumerate}
    \item \textbf{Skeleton Retargeting and Prefix Conditioning:} We retarget the 22-joint HumanML3D seed history to the 27-joint \textsc{CoreSkeleton27} layout. The first 8 frames of the 10-frame seed are given as ARDY's native motion history, and the last two seed frames are imposed as position-only soft conditioning, with wrist, hand, and thumb joints excluded from the soft constraints.
    \item \rev{\textbf{Future Contact Inpainting:} Target contact schedules are converted into foot-height targets and imposed through ARDY's native joint-position conditioning. During scheduled stance intervals, the ankle and toe heights are constrained to $0.06\,\text{m}$ and $0.00\,\text{m}$ in the inpainting mask, and during each swing phase $0.08\sin^2(\pi\,\text{phase})\,\text{m}$ is added to the stance height.}
    \item \textbf{Sampling and Post-Processing:} We sample motions using 10 denoising steps with text guidance weight $2.0$ and spatial constraint guidance weight $0.9$. Decoded motions are post-processed with ARDY's official motion-correction procedure (\texttt{post\_process\_motion}), which uses the contact predicted internally by ARDY.
    \item \textbf{Evaluation in the Native CORE Space:} Generated motion and retargeted ground truth remain in the native 27-joint \textsc{CoreSkeleton27} representation.
    Contact timing, stance slip, and floor penetration are measured directly from ARDY's native foot joints.
    Only for RA-LB MPJPE, we select the 22 semantically corresponding joints from both CORE trajectories and apply the same translation-and-heading alignment used for the other methods.
\end{enumerate}

\subsection{Unified Physical Contact Decision Criteria}
\label{sec:supp_unified_contact}
To ensure completely fair and unbiased comparison across all evaluated models (StepControl, DART, ARDY, and Ground Truth), we do not rely on model-internal predicted contact flags or arbitrary heuristic flags. Instead, we extract realized contact states directly from decoded 3D kinematic joint trajectories using a kinematic decision rule, where $v$ denotes the per-frame displacement of a joint:
\begin{equation}
\begin{aligned}
    \hat{c}_{t,\text{foot}} = \mathbb{I}\Big( & \min(v_{t,\text{ankle}}^2, v_{t,\text{toe}}^2) < 0.002\,\text{m}^2/\text{frame}^2 \\
    & \wedge y_{t,\text{ankle}} \le 0.14\,\text{m} \;\wedge\; y_{t,\text{toe}} \le 0.08\,\text{m} \Big).
\end{aligned}
\end{equation}
This rule requires both (i) low per-frame foot displacement ($\|v\| < 0.0447\,\text{m/frame}$, about $0.89\,\text{m/s}$ at 20\,fps) and (ii) geometric proximity to the ground plane for both heel and toe joints.

\section{Inference Runtime and Generation Throughput}
\label{sec:supp_runtime}

\begin{table}[t]
\caption{\textbf{Inference runtime and generation throughput comparison for 20-second (400-frame) rollouts on a single NVIDIA RTX 4090 GPU.} StepControl achieves real-time throughput ($>130$\,fps) via direct single-pass generation.}
\centering
\small
\resizebox{\linewidth}{!}{
\begin{tabular}{lccc}
\toprule
\textbf{Method} & \textbf{Per-Frame} & \textbf{20\,s Rollout} & \textbf{Throughput} \\
\midrule
DART~\cite{dart} (50-step Opt.) & $\approx 2267\,\text{ms}$ & $\approx 907\,\text{s}$ & $0.44\,\text{fps}$ \\
ARDY~\cite{ardy} (10-step Diff.) & $\approx 75.0\,\text{ms}$ & $\approx 30.0\,\text{s}$ & $13.3\,\text{fps}$ \\
\midrule
\textbf{StepControl (Ours)} & $\mathbf{7.5\,\text{ms}}$ & $\mathbf{2.9\,\text{s}}$ & $\mathbf{133.3\,\text{fps}}$ \\
\bottomrule
\end{tabular}
}
\label{tab:supp_runtime}
\end{table}

In~\Cref{tab:supp_runtime}, we report the computational efficiency and latency of each method for generating a 20-second (400-frame at 20\,fps) continuous motion rollout on a single NVIDIA RTX 4090 GPU:
\begin{itemize}
    \item \textbf{DART~\cite{dart}:} Requires 50 optimizer iterations per primitive ($F=8$ frames), i.e., 2,500 iterations for a 20-second sequence, each involving diffusion sampling and decoding of multiple primitives. This test-time optimization takes $\approx 15.1$ minutes per 20-second sequence ($0.44\,\text{fps}$).
    \item \textbf{ARDY~\cite{ardy}:} Evaluates 10 diffusion denoising steps with text and inpainting guidance over 40-frame sliding windows, taking $\approx 30.0$ seconds ($13.3\,\text{fps}$).
    \item \textbf{StepControl (Ours):} Generates each future frame in a single feed-forward pass via progressive DDIM refinement ($1$ step per frame shift), achieving $\mathbf{7.5\,\text{ms}}$ per frame and $\mathbf{133.3\,\text{fps}}$ throughput ($6.7\times$ faster than real-time playback).
    \item \textbf{StepControl (+Opt.):} The unconstrained planar root correction executes in $< 1\,\text{ms}$ in a single linear pass ($\mathcal{O}(T)$), preserving identical real-time throughput.
\end{itemize}

\section{Additional Ablation Studies}
\label{sec:supp_ablations}

\begin{table*}[t]
\caption{\textbf{Ablation on loss formulations (LAFAN1 test benchmark).} All variants use the same mixture noise schedule, pretraining initialization, fine-tuning protocol, evaluation segments, and random seed.}
\centering
\small
\resizebox{0.95\linewidth}{!}{
\begin{tabular}{lccccc}
\toprule
\textbf{Loss Formulation} & \textbf{Bal. Acc. (\%)} $\uparrow$ & \textbf{Tr. F1 (\%)} $\uparrow$ & \textbf{Slip (mm/f)} $\downarrow$ & \textbf{RA-LB (mm)} $\downarrow$ & \textbf{Bo. Jerk (mm/f$^3$)} $\downarrow$ \\
\midrule
Only Motion Reconstruction Loss ($\mathcal{L}_{\text{mse}}$) & 73.87 & 21.60 & 22.40 & 204.35 & 71.45 \\
w/o Stance Velocity Loss ($\mathcal{L}_{\text{stance}}$) & 76.84 & 32.27 & 19.93 & 201.03 & 70.94 \\
w/o Foot Height Loss ($\mathcal{L}_{\text{height}}$) & 78.02 & 33.30 & \textbf{15.39} & \textbf{180.55} & 70.68 \\
w/o 4D Contact MSE Loss ($\mathcal{L}_{\text{contact4}}$) & 79.03 & 34.21 & 15.80 & 181.17 & \textbf{70.28} \\
\midrule
\textbf{Full Multi-Task Loss (StepControl)} & \textbf{79.30} & \textbf{39.58} & 15.54 & 181.64 & 70.54 \\
\bottomrule
\end{tabular}
}
\label{tab:supp_loss_ablation}
\end{table*}

\begin{table*}[t]
\caption{\textbf{Ablation on classifier-free guidance (CFG) scale $s$ (LAFAN1 test benchmark).} Increasing $s$ enhances temporal contact adherence and transition precision.}
\centering
\small
\resizebox{0.95\linewidth}{!}{
\begin{tabular}{lccccc}
\toprule
\textbf{CFG Guidance Scale} & \textbf{Bal. Acc. (\%)} $\uparrow$ & \textbf{Tr. F1 (\%)} $\uparrow$ & \textbf{Slip (mm/f)} $\downarrow$ & \textbf{RA-LB (mm)} $\downarrow$ & \textbf{Bo. Jerk (mm/f$^3$)} $\downarrow$ \\
\midrule
$s = 0.0$ (Unconditional Baseline) & 52.08 & 3.90 & 38.36 & 208.98 & \textbf{69.12} \\
$s = 1.0$ (Standard Conditional Prior) & 74.67 & 24.52 & 16.91 & 183.16 & 69.49 \\
\midrule
\textbf{$s = 2.0$ (Default StepControl)} & \textbf{79.30} & \textbf{39.58} & \textbf{15.54} & \textbf{181.64} & 70.54 \\
\bottomrule
\end{tabular}
}
\label{tab:supp_cfg_ablation}
\end{table*}

\subsection{Ablation on Training Loss Formulations}
\label{sec:supp_loss_ablation}

In~\Cref{tab:supp_loss_ablation}, we evaluate the necessity of each constituent term in the multi-task training objective:
\begin{equation}
\begin{aligned}
    \mathcal{L} = {} & \mathcal{L}_{\text{mse}} + \lambda_{\text{p}}\mathcal{L}_{\text{prefix}} + \lambda_{\text{vel}}\mathcal{L}_{\text{stance}} \\
    & + \lambda_{\text{h}}\mathcal{L}_{\text{height}} + \lambda_{\text{c}}\mathcal{L}_{\text{contact4}}.
\end{aligned}
\end{equation}
\begin{itemize}
    \item \textbf{Motion-Only Supervision ($\mathcal{L}_{\text{mse}}$):} Training with only the standardized feature MSE (future and prefix frames, without the geometric and contact losses) substantially degrades contact timing ($\text{Tr. F1} = 21.60\%$), increases stance slip to $22.40\,\text{mm/f}$, and raises lower-body error to $204.35\,\text{mm}$.
    \item \textbf{Role of Stance Velocity Loss ($\mathcal{L}_{\text{stance}}$):} Omitting $\mathcal{L}_{\text{stance}}$ increases stance foot slip from $15.54$ to $19.93\,\text{mm/f}$ and reduces transition F1 from $39.58\%$ to $32.27\%$.
    \item \textbf{Role of Foot Height Loss ($\mathcal{L}_{\text{height}}$):} Removing explicit foot-height supervision reduces transition F1 from $39.58\%$ to $33.30\%$, although it does not degrade the root-aligned pose or slip metrics in this evaluation.
    \item \textbf{Role of 4D Contact MSE Loss ($\mathcal{L}_{\text{contact4}}$):} Removing auxiliary contact reconstruction lowers transition F1 from $39.58\%$ to $34.21\%$, showing that the discrete contact channels provide complementary timing supervision.
    \item \textbf{Boundary Smoothness:} Boundary jerk varies only from $70.28$ to $71.45\,\text{mm/f}^3$ across loss variants. Removing individual geometric losses has a limited effect on boundary jerk in this evaluation.
\end{itemize}

\subsection{Ablation on Classifier-Free Guidance Scale}
\label{sec:supp_cfg_ablation}

\Cref{tab:supp_cfg_ablation} investigates the effect of the classifier-free guidance scale $s$ during sampling:
\begin{itemize}
    \item At $s = 0.0$ (unconditional generation), the model ignores the contact schedule entirely, resulting in near-random contact synchronization ($\text{Tr. F1} = 3.90\%$, $\text{Bal. Acc} = 52.08\%$) and high foot slip ($38.36\,\text{mm/f}$).
    \item At $s = 1.0$ (standard conditional sampling), the model begins to follow the target schedule ($\text{Tr. F1} = 24.52\%$, $\text{Bal. Acc} = 74.67\%$).
    \item Increasing guidance to $s = 2.0$ (our default setting) further strengthens contact adherence to $\text{Tr. F1} = 39.58\%$ and $\text{Bal. Acc} = 79.30\%$, while reducing stance foot slip from $16.91$ to $15.54\,\text{mm/f}$ and lowering lower-body kinematic error to $181.64\,\text{mm}$.
\end{itemize}

\subsection{Rollout Strategy Comparison}
\label{sec:supp_rollout_ablation}

\begin{table*}[t]
\caption{\rev{\textbf{Rollout strategy comparison on the LAFAN1 test benchmark.}}}
\centering
\small
\resizebox{0.85\linewidth}{!}{
\begin{tabular}{lccccc}
\toprule
\textbf{Inference} & \textbf{Bal. Acc. (\%)} $\uparrow$ & \textbf{Tr. F1 (\%)} $\uparrow$ & \textbf{Slip (mm/f)} $\downarrow$ & \textbf{RA-LB (mm)} $\downarrow$ & \textbf{Bo. Jerk (mm/f$^3$)} $\downarrow$ \\
\midrule
Chunk AR & 76.32 & 30.75 & 19.96 & 203.63 & 75.59 \\
\textbf{Progressive (Ours)} & \textbf{79.30} & \textbf{39.58} & \textbf{15.54} & \textbf{181.64} & \textbf{70.54} \\
\bottomrule
\end{tabular}
}
\label{tab:supp_rollout}
\end{table*}

We compare progressive rollout with chunk-wise autoregressive rollout, which generates 20 future frames at once from 10 history frames and uses the last 10 of the generated frames as the history for the next chunk. Both use the same mixture-trained model, history and future lengths, CFG scale, and LAFAN1 test sequences, without post-processing. Progressive rollout outperforms chunk-wise rollout across all metrics (\Cref{tab:supp_rollout}), and the same trend holds for transition F1 at the 20-second horizon.

\subsection{Analysis by Contact-Transition Rate}
\label{sec:supp_rate_analysis}

\begin{table}[t]
\caption{\rev{\textbf{Transition F1 (\%) by contact-transition rate.}}}
\centering
\small
\resizebox{\linewidth}{!}{
\begin{tabular}{lccc}
\toprule
 & \textbf{Low} & \textbf{Mid} & \textbf{High} \\
\midrule
Transition rate (events/s) & 1.2--3.2 & 3.4--4.0 & 4.4--8.0 \\
Windows & 20 & 16 & 16 \\
Target events & 253 & 300 & 460 \\
\midrule
DART & 17.82 & 15.93 & 16.88 \\
ARDY & 13.42 & 16.64 & 18.54 \\
Raw 2D Contact & 26.55 & 33.34 & 25.65 \\
Delta 5D Contact & 24.73 & 33.85 & 35.07 \\
\textbf{Event 9D Contact (Ours)} & \textbf{37.56} & \textbf{48.86} & \textbf{41.46} \\
\bottomrule
\end{tabular}
}
\label{tab:supp_rate}
\end{table}

The 20-second rollouts are split into non-overlapping 5-second windows and grouped by the number of contact transitions per second. Events are pooled within each group, and onset and release F1 are averaged. Event 9D contact conditioning yields higher transition F1 than the raw 2D contact input in every rate group, and remains ahead of both spatially guided baselines (\Cref{tab:supp_rate}).

\section{Unconstrained Planar Stance Foot Correction (StepControl +Opt.)}
\label{sec:supp_foot_opt}

In the main paper, we report both raw generations (\textbf{StepControl}) and an optional post-processed variant (\textbf{StepControl +Opt.}). Here, we formally describe the planar stance-foot correction algorithm.

\paragraph{Motivation and Core Principle}
Unlike prior works that rely on heavy test-time gradient descent or inverse kinematics (IK) solvers that can distort natural knee flexion and limb articulation, our correction operates purely at the level of the global horizontal root trajectory ($x$-$z$ ground plane). It directly cancels residual stance-phase horizontal foot drift without altering joint angles, vertical foot clearances, or local body poses.

\paragraph{Formulation and Algorithm}
Let $\hat{\mathbf{J}}_{1:T} \in \mathbb{R}^{T \times 22 \times 3}$ denote the synthesized 3D joint positions and $\mathbf{c}_{1:T} \in \{0, 1\}^{T \times 2}$ denote the prescribed left/right foot contact schedule. At each frame $t \in [S+1, T]$:
\begin{enumerate}
    \item \textbf{Persistent Stance Identification:} We identify active feet in persistent stance across consecutive frames:
    \begin{equation}
        \mathcal{A}_t = \{ f \in \{\text{L}, \text{R}\} \mid c_{t-1, f} = 1 \wedge c_{t, f} = 1 \}.
    \end{equation}
    \item \textbf{Horizontal Stance Displacement:} For each active foot $f \in \mathcal{A}_t$, we measure the horizontal displacement between successive frames across its constituent ankle and toe joints $\mathcal{F}_f$:
    \begin{equation}
        \Delta \mathbf{p}_{t, f}^{xz} = \arg\min_{\mathbf{d} \in \mathcal{D}_{t, f}} \|\mathbf{d}\|_2,
    \end{equation}
    where $\mathcal{D}_{t, f} = \{\hat{\mathbf{J}}_{t, j}^{xz} - \hat{\mathbf{J}}_{t-1, j}^{xz} \mid j \in \mathcal{F}_f\}$.
    \item \textbf{Accumulated Planar Root Offset:} The cumulative horizontal correction $\mathbf{r}_t^{\text{corr}} \in \mathbb{R}^2$ is updated recursively:
    \begin{equation}
        \mathbf{r}_t^{\text{corr}} = \begin{cases}
        \mathbf{r}_{t-1}^{\text{corr}} - \frac{1}{|\mathcal{A}_t|} \sum_{f \in \mathcal{A}_t} \Delta \mathbf{p}_{t, f}^{xz}, & \text{if } \mathcal{A}_t \neq \emptyset, \\
        \mathbf{r}_{t-1}^{\text{corr}}, & \text{otherwise}.
        \end{cases}
    \end{equation}
    \item \textbf{Global Translation Update:} The computed offset is applied uniformly to all joint coordinates along the ground plane:
    \begin{equation}
        \hat{\mathbf{J}}_{t, j}^{xz} \leftarrow \hat{\mathbf{J}}_{t, j}^{xz} + \mathbf{r}_t^{\text{corr}}, \quad \forall j \in \{1, \dots, 22\}.
    \end{equation}
\end{enumerate}

\section{Network Architecture and Training Details}
\label{sec:supp_arch_training}

\paragraph{Transformer Architecture}
The contact-aware motion denoiser is built upon a spatio-temporal Transformer architecture:
\begin{itemize}
    \item \textbf{Window Size:} $W = 30$ frames ($1.5$\,s at 20\,fps), divided into $S = 10$ history frames ($0.5$\,s) and $F = 20$ future frames ($1.0$\,s).
    \item \textbf{Transformer Layers:} 8 transformer encoder layers.
    \item \textbf{Hidden Dimension:} $d = 512$, with feed-forward inner dimension $d_{\text{ff}} = 2048$ and dropout $0.1$.
    \item \textbf{Attention Heads:} 8 multi-head self-attention heads with head dimension $d_k = 64$.
    \item \textbf{Cross-Attention Blocks:} Per-layer multi-head cross-attention over contact tokens produced from the 9D event-aware features by a two-layer MLP with SiLU ($\mathbf{E}_{1:W} \to \mathbf{Z} \in \mathbb{R}^{W \times 512}$, $9 \to 512 \to 512$).
    \item \textbf{Activation and Normalization:} GELU activations in the feed-forward layers with pre-layer normalization (Pre-LN); the contact embedding MLP uses SiLU.
\end{itemize}

\paragraph{Diffusion Hyperparameters and Noise Schedule}
\begin{itemize}
    \item \textbf{Total Diffusion Steps:} $K = 100$ discrete noise levels ($k \in \{0, \dots, 99\}$).
    \item \textbf{Variance Schedule:} Cosine variance schedule with offset $0.008$ and maximum $\beta$ of $0.999$.
    \item \textbf{Progressive Rollout Slope:} $\mathbf{k}_{\text{roll}} = [4, 9, 14, \dots, 99]$ across future frames $t \in [S+1, W]$ with stride $\Delta k = 5$.
    \item \textbf{Training Noise Distribution:} Stochastic mixture sampling with $70\%$ probability assigned to the structured slope $\mathbf{k}_{\text{roll}}$ and $30\%$ to uniform IID noise $k_t \sim \mathcal{U}\{0, 99\}$ in both training stages.
\end{itemize}

\paragraph{Optimization and Training Setup}
\begin{itemize}
    \item \textbf{Optimizer:} AdamW with $\beta_1 = 0.9$, $\beta_2 = 0.999$, and weight decay $0.01$. Gradient clipping is $1.0$ and the EMA decay is $0.999$.
    \item \textbf{Learning Rate:} $10^{-4}$ for HumanML3D pretraining and $5 \times 10^{-5}$ for LAFAN1 fine-tuning. Fine-tuning is initialized from the pretrained EMA weights with a newly initialized optimizer.
    \item \textbf{Batch Size and Schedule:} Batch size 128 for $20$ epochs in each stage ($3{,}660$ updates for pretraining and $5{,}000$ for fine-tuning), with training seed $2026$.
    \item \textbf{Loss Weights:} $\lambda_{\text{vel}} = 20$, $\lambda_{\text{h}} = 20$, and $\lambda_{\text{c}} = 0.1$, and $\lambda_{\text{p}} = 0.1$. Conditional and unconditional branches are weighted equally.
\end{itemize}

\section{Evaluation Metric Definitions}
\label{sec:supp_metrics}

Let $\mathbf{J}_t \in \mathbb{R}^{22 \times 3}$ and $\hat{\mathbf{J}}_t \in \mathbb{R}^{22 \times 3}$ denote ground-truth and generated 3D joint positions at frame $t$, respectively. Let $\mathbf{c}_t = [c_{t,\text{L}}, c_{t,\text{R}}] \in \{0, 1\}^2$ denote the target binary contact schedule and $\hat{\mathbf{c}}_t \in \{0, 1\}^2$ denote the realized foot-contact state extracted from generated foot kinematics using the kinematic contact rule in \Cref{sec:supp_unified_contact}.
Joint positions are measured in meters; the factor $10^3$ converts distance-based metrics to the reporting units given in the paragraph headings below.

\paragraph{Balanced Contact Accuracy}
Balanced Contact Accuracy is the unweighted mean of stance recall ($\text{Rec}_{\text{stance}}$) and swing recall ($\text{Rec}_{\text{swing}}$) across all evaluated frames $t \in [S+1, T]$:
\begin{equation}
\begin{aligned}
    \text{Bal. Acc.} = \frac{100\%}{2}\Bigg( & \frac{\sum_{t} \mathbb{I}(\hat{c}_t = 1 \wedge c_t = 1)}{\sum_t \mathbb{I}(c_t = 1)} \\
    & + \frac{\sum_{t} \mathbb{I}(\hat{c}_t = 0 \wedge c_t = 0)}{\sum_t \mathbb{I}(c_t = 0)} \Bigg).
\end{aligned}
\end{equation}

\paragraph{Contact Transition F1 (Tr. F1)}
Discrete onset ($0 \to 1$) and release ($1 \to 0$) transition events are extracted via temporal difference:
\begin{equation}
\begin{aligned}
    e_{t,\text{onset}} &= \mathbb{I}(c_t = 1 \wedge c_{t-1} = 0), \\
    e_{t,\text{release}} &= \mathbb{I}(c_t = 0 \wedge c_{t-1} = 1).
\end{aligned}
\end{equation}
Precision, recall, and harmonic mean F1-scores ($\text{F1}_{\text{onset}}$ and $\text{F1}_{\text{release}}$) are computed against the target schedule events with a tolerance window of 0 frames. The reported transition F1 is:
\begin{equation}
    \text{Tr. F1} = \frac{1}{2}\left( \text{F1}_{\text{onset}} + \text{F1}_{\text{release}} \right) \times 100\%.
\end{equation}

\paragraph{Stance Foot Slip ($\mathrm{mm/frame}$)}
Measures the mean 3D displacement of a foot between consecutive frames in which the foot is in stance in both frames. For each foot, the smaller of the ankle and toe displacements is used:
\begin{equation}
    \text{Slip} = 10^3 \frac{\sum_{f}\sum_{t \in \mathcal{S}_f} \min_{j \in \mathcal{F}_f} \|\hat{\mathbf{J}}_{t+1,j} - \hat{\mathbf{J}}_{t,j}\|_2}{\sum_{f} |\mathcal{S}_f|},
\end{equation}
where $\mathcal{S}_f = \{t \in [S+1, T-1] \mid c_{t,f} = 1 \wedge c_{t+1,f} = 1\}$ and $\mathcal{F}_f$ denotes the ankle and toe joints of foot $f$. Slip is pooled over all valid foot-frames within a sequence, and the reported value is the mean over sequences.

\paragraph{Floor Penetration ($\mathrm{mm}$)}
Quantifies vertical violations below the floor ground plane ($y = 0$):
\begin{equation}
    \text{Pen.} = \frac{10^3}{(T-S)|\mathcal{F}|} \sum_{t=S+1}^T \sum_{j \in \mathcal{F}} \max(0, -\hat{y}_{t,j}).
\end{equation}

\paragraph{Root-Aligned Lower-Body MPJPE (RA-LB, $\mathrm{mm}$)}
Evaluates lower-limb pose tracking independently of accumulated global translation and yaw drift:
\begin{equation}
    \text{RA-LB} = \frac{10^3}{(T-S)|\mathcal{L}|} \sum_{t=S+1}^T \sum_{j \in \mathcal{L}} \|\mathcal{T}_{\text{align}}(\hat{\mathbf{J}}_{t,j}) - \mathbf{J}_{t,j}\|_2,
\end{equation}
where $\mathcal{L}$ indexes the lower-body joints (pelvis, hips, knees, ankles, toes) and $\mathcal{T}_{\text{align}}$ aligns the generated pose to the ground-truth root position along all three axes and facing heading angle at each frame.

\paragraph{Boundary Jerk ($\mathrm{mm/frame^3}$)}
Quantifies physical smoothness and absence of popping shocks at the observed prefix to generated motion transition ($t = S \to S+1$):
\begin{equation}
\begin{aligned}
    \text{Bo. Jerk} = \frac{10^3}{21} \sum_{j=2}^{22} \big\| & (\hat{\mathbf{p}}_{S+1,j} - 2\hat{\mathbf{p}}_{S,j} + \hat{\mathbf{p}}_{S-1,j}) \\
    & - (\hat{\mathbf{p}}_{S,j} - 2\hat{\mathbf{p}}_{S-1,j} + \hat{\mathbf{p}}_{S-2,j}) \big\|_2,
\end{aligned}
\end{equation}
where $\hat{\mathbf{p}}_{t,j} = \hat{\mathbf{J}}_{t,j} - \hat{\mathbf{J}}_{t,1}$ denotes root-relative joint positions.

\end{document}